\documentclass[letterpaper]{article} % DO NOT CHANGE THIS
\usepackage{aaai2026}
\nocopyright
\usepackage{times}  % DO NOT CHANGE THIS
\usepackage{helvet}  % DO NOT CHANGE THIS
\usepackage{courier}  % DO NOT CHANGE THIS
\usepackage[hyphens]{url}  % DO NOT CHANGE THIS
\usepackage{graphicx} % DO NOT CHANGE THIS
\usepackage{natbib}  % DO NOT CHANGE THIS AND DO NOT ADD ANY OPTIONS TO IT
\usepackage{caption} % DO NOT CHANGE THIS AND DO NOT ADD ANY OPTIONS TO IT
\usepackage{amsmath}
\usepackage{algpseudocode}
\usepackage{xspace}
\usepackage{amssymb}
\usepackage{booktabs}
\usepackage{cleveref}
\usepackage{makecell}
\usepackage{arydshln}
\usepackage{algorithm}
\usepackage{newfloat}
\usepackage{listings}

\usepackage{makecell}      % preamble

\newcommand{\m}[2]{\makecell[l]{#1\\#2}}

\DeclareCaptionStyle{ruled}{labelfont=normalfont,labelsep=colon,strut=off} % DO NOT CHANGE THIS
\floatstyle{ruled}
\newfloat{listing}{tb}{lst}{}
\floatname{listing}{Listing}
\title{Streaming Generation of Co-Speech Gestures via Accelerated Rolling Diffusion}
\author {
    Evgeniia Vu\textsuperscript{\rm 1}\equalcontrib,
    Andrei Boiarov\textsuperscript{\rm 2}\equalcontrib,
    Dmitry Vetrov\textsuperscript{\rm 1}
}
\affiliations {
    \textsuperscript{\rm 1}Constructor University, Bremen, Germany\\
    \textsuperscript{\rm 2}Constructor Tech, Sofia, Bulgaria\\
    andrei.boiarov@constructor.tech
}
\begin{document}

\maketitle

% \renewcommand{\thefootnote}{\fnsymbol{footnote}}
% \footnotetext[1]{Equal contribution. Correspondence to: andrei.boiarov@constructor.tech}

\begin{abstract}
Generating co-speech gestures in real time requires both temporal coherence and efficient sampling. We introduce a novel framework for streaming gesture generation that extends Rolling Diffusion models with structured progressive noise scheduling, enabling seamless long-sequence motion synthesis while preserving realism and diversity. Our framework is universally compatible with existing diffusion-based gesture generation model, transforming them into streaming methods capable of continuous generation without requiring post-processing. We evaluate our framework on ZEGGS and BEAT, strong benchmarks for real-world applicability. Applied to state-of-the-art baselines on both datasets, it consistently outperforms them, demonstrating its effectiveness as a generalizable and efficient solution for real-time co-speech gesture synthesis. We further propose Rolling Diffusion Ladder Acceleration (RDLA), a new approach  that employs a ladder-based noise scheduling strategy to simultaneously denoise multiple frames. This significantly improves sampling efficiency while maintaining motion consistency, achieving up to a 4× speedup with high visual fidelity and temporal coherence in our experiments. Comprehensive user studies further validate our framework’s ability to generate realistic, diverse gestures closely synchronized with the audio input.
\end{abstract}


\begin{links}
    \link{Code}{https://github.com/andrewbo29/co-speech-gestures-rolling-diffusion}
\end{links}

\section{Introduction}

% Co-speech gestures significantly enhance non-verbal communication by reinforcing spoken content, crucially contributing to realism in virtual avatars, video conferencing, gaming, and interactive embodied AI applications~\cite{nyatsanga2023comprehensive}. Real-time generation of these gestures, or streaming generation, is essential in scenarios like virtual assistants, gaming, and telepresence systems. Recent approaches predominantly leverage data-driven deep learning methods to produce realistic motion sequences. While prior frameworks have shown success, diffusion-based models have recently emerged as particularly effective due to their exceptional realism and diversity. Notably, methods such as DiffuseStyleGesture~\cite{yang2023diffusestylegesture} use cross-local and self-attention for gesture-speech synchronization, Taming Diffusion~\cite{zhu2023tamingdiffusionmodelsaudiodriven} frames gesture synthesis as audio-conditioned diffusion with transformer guidance for temporally coherent motion, both inherently apply direct conditioning on seed poses to ensure smooth transitions.

Co-speech gestures enhance non-verbal communication by reinforcing spoken content, and are essential for realism in virtual avatars, video conferencing, gaming, and embodied AI~\cite{nyatsanga2023comprehensive}. Real-time, or streaming, generation is particularly important for interactive settings such as virtual assistants, gaming, and telepresence systems. Recent work largely relies on data-driven deep learning to produce realistic motion, with diffusion-based models standing out for their strong realism and diversity. Approaches such as DiffuseStyleGesture~\cite{yang2023diffusestylegesture}, which uses cross-local and self-attention for gesture–speech synchronization, and Taming Diffusion~\cite{zhu2023tamingdiffusionmodelsaudiodriven}, which applies audio-conditioned diffusion with transformer guidance, both rely on seed-pose conditioning to ensure smooth transitions and temporal coherence.

Despite recent progress, diffusion-based methods still struggle in real-time settings. To manage long sequences, many models generate fixed-length chunks or incremental extensions. PersonaGestor~\cite{zhang2024speechdrivenpersonalizedgesturesynthetics} uses fuzzy feature inference yet still stitches chunked outputs, while DiffSHEG~\cite{chen2024diffsheg} depends on incremental outpainting. Such non-streaming strategies often cause visual discontinuities and add latency due to post-processing. Methods that condition on previous frames improve continuity but incur substantial computational overhead, limiting real-time use (see Supplementary Material).

% A promising direction for sequential generation is the Rolling Diffusion framework~\cite{ruhe2024rollingdiffusionmodels}, which extends traditional diffusion to an autoregressive process. By iteratively generating new data conditioned on previous outputs, this approach can enhance temporal consistency for long sequences. However, adapting this framework for a practical, real-time application like co-speech gesture generation remains an open challenge, particularly concerning the computational demands of the autoregressive loop.

% A promising direction for sequential generation is the Rolling Diffusion framework~\cite{ruhe2024rollingdiffusionmodels}, which extends diffusion models into an autoregressive process. By iteratively generating new data conditioned on previous outputs, it improves temporal consistency for long sequences. However, applying this framework to real-time co-speech gesture generation remains challenging, largely due to the computational cost of the autoregressive loop.

A promising alternative is the Rolling Diffusion framework~\cite{ruhe2024rollingdiffusionmodels}, which turns diffusion models into an autoregressive process and improves temporal consistency for long sequences. However, applying it to real-time co-speech gesture generation remains difficult, primarily because the autoregressive loop is computationally expensive.

To address these challenges, we introduce a novel Rolling Diffusion based framework integrating diffusion models with real-time capabilities for co-speech gesture generation. By employing a structured ladder-based noise scheduling strategy, our Rolling Diffusion Ladder Acceleration (RDLA) approach simultaneously denoises multiple frames, significantly improving sampling efficiency. This achieves generation speeds of up to 200 FPS without compromising visual fidelity or temporal coherence. Extensive experiments on the ZEGGS~\cite{ghorbani2023zeroeggs} and BEAT~\cite{liu2022beatlargescalesemanticemotional} benchmarks confirm our method’s effectiveness for realistic, diverse gesture generation in streaming contexts. In summary, our key contributions are:
\begin{enumerate}
\item We are the first, to our knowledge, to successfully adapt a rolling diffusion framework to a practical application, specifically demonstrating its effectiveness in real-time co-speech gesture generation.
\item We propose a universal framework for converting any diffusion-based gesture generation approach into a real-time streaming model without requiring post-processing.
\item We provide comprehensive evaluations on standard benchmarks and user studies, showing that our method consistently outperforms baselines across the benchmarks, demonstrating its effectiveness as a generalizable solution for real-time, high-fidelity co-speech gesture synthesis.
\item We introduce RDLA, substantially increasing inference speed with minimal impact on gesture quality.
\end{enumerate}

\section{Proposed Approach}
\subsection{Rolling Diffusion Models}
Diffusion models~\cite{song2020generativemodelingestimatinggradients, ho2020denoising} consist of a forward (diffusion) process and reverse process. The forward process gradually adds Gaussian noise to a data sample \( x^0 \) over \( T \) steps such that \(x^T \sim \mathcal{N}(0, 1)\) and each transition is defined as:

\begin{equation}
q(x^t \mid x^{t-1}) = \mathcal{N}(x^t; \sqrt{1 - \beta^t} x^{t-1}, \beta^t I)
\end{equation}
where \( \beta^t \) is a variance schedule controlling noise intensity. The marginal distribution after \( t \) steps is:  

\begin{equation}
q(x^t \mid x^0) = \mathcal{N}(x^t; \sqrt{\bar{\alpha}^t} x^0, (1 - \bar{\alpha}^t) I)
\end{equation}
where \( \alpha^t = 1 - \beta^t \) and \( \bar{\alpha}^t = \prod_{s=1}^{t} \alpha^s \).  \\
The reverse process learns to denoise \( x^t \) using a neural network \( f_\theta(x^t, t) = \hat{x}\). The model is trained by minimizing the objective:
$\mathcal{L}(\theta) := \mathbb{E}_{t, x^0} 
\left[ a(t) \|x^0 - \hat{x}\|^2 \right]$, where \(a(t)\) is a weighting function that can be specified to control the importance of different timesteps during training.
This objective encourages \( f_\theta(x^t, t) \) to accurately estimate the initial signal of data sample.  

Rolling Diffusion Models (RDMs) \cite{ruhe2024rollingdiffusionmodels} introduce a modification to standard diffusion models by incorporating a progressive corruption process along the temporal axis, making them particularly well-suited for sequential data \( \mathbf{X} = \{x^0_k\}_{k = 0}^{L - 1} \), where \( x_l\) is $k$-th element of sequence. Unlike standard diffusion models that apply noise uniformly across all frames, RDMs operate with a rolling window of size $N$. In this setup, the noise level gradually increases from the first to the last frame in the window, enabling a seamless transition. In our implementation, we discretize time which indicates the noise level using \( t \in \{0,\ldots, T\} \) instead of the continuous range \( t \in [0,1] \), where \( T = 1000 \) represents the total number of noise steps. Since the generation window size \( N \) is typically much smaller than \( T \), the noise level difference between adjacent frames is greater than 1.  We define this difference as a step \( s \), calculated as
\(s = \frac{T}{N}\). 
To ensure uniform noise distribution across frames, we select the generation window as a divisor of \( T \), ensuring consistent step sizes and preventing uneven noise application. This structured noise scheduling allows for a more controlled and stable generation process, improving the overall quality of generated sequences. Then the rolling window takes the form  \(\mathbf{x}^{t_0}_j = \{x_{j + n}^{t_n}\}_{n=0}^{N-1}\), where $t_0\in\{1,\ldots,s\}$ is a noise level of the first frame in the window, $t_n=t_0+ns$ is a noise level of $n^{th}$ frame. \footnote{Since noise levels of all frames within a rolling window are deterministic functions of $t_0$ we may use this index to define the noise levels of the entire rolling window $\mathbf{x}_j^{t_0}$.} 

In the forward process, noise is applied progressively as:
\begin{equation}\label{eq:rdm_q}
q(\mathbf{x}^{t_0}_j | \mathbf{x}_j) = \prod_{n=0}^{N-1} \mathcal{N}(x^{t_n}_{j + n} | \alpha^{t_n} x_{j+n}^0, (1-\alpha^{t_n})^2 I)
\end{equation}
During training, the model \( f_\theta(\mathbf{x}_j^{t_0}, t) = \hat{\mathbf{x}}\) processes only the frames within the rolling window. The parametrized reverse process $p_{\theta}(\mathbf{x}^{t - 1}_j | \mathbf{x}^{t}_j)$ is defined as:
\begin{equation}\label{eq:rdm_p}
p_{\theta}(\mathbf{x}^{t_0 - 1}_j | \mathbf{x}^{t_0}_j) = \prod_{n=0}^{N-1}q(x_{j + n}^{t_n - 1} | x_{j + n}^{t_n}, \hat{x}_{j + n})
\end{equation}

\begin{figure}[!t]
    \centering
    \includegraphics[width=0.45\textwidth]{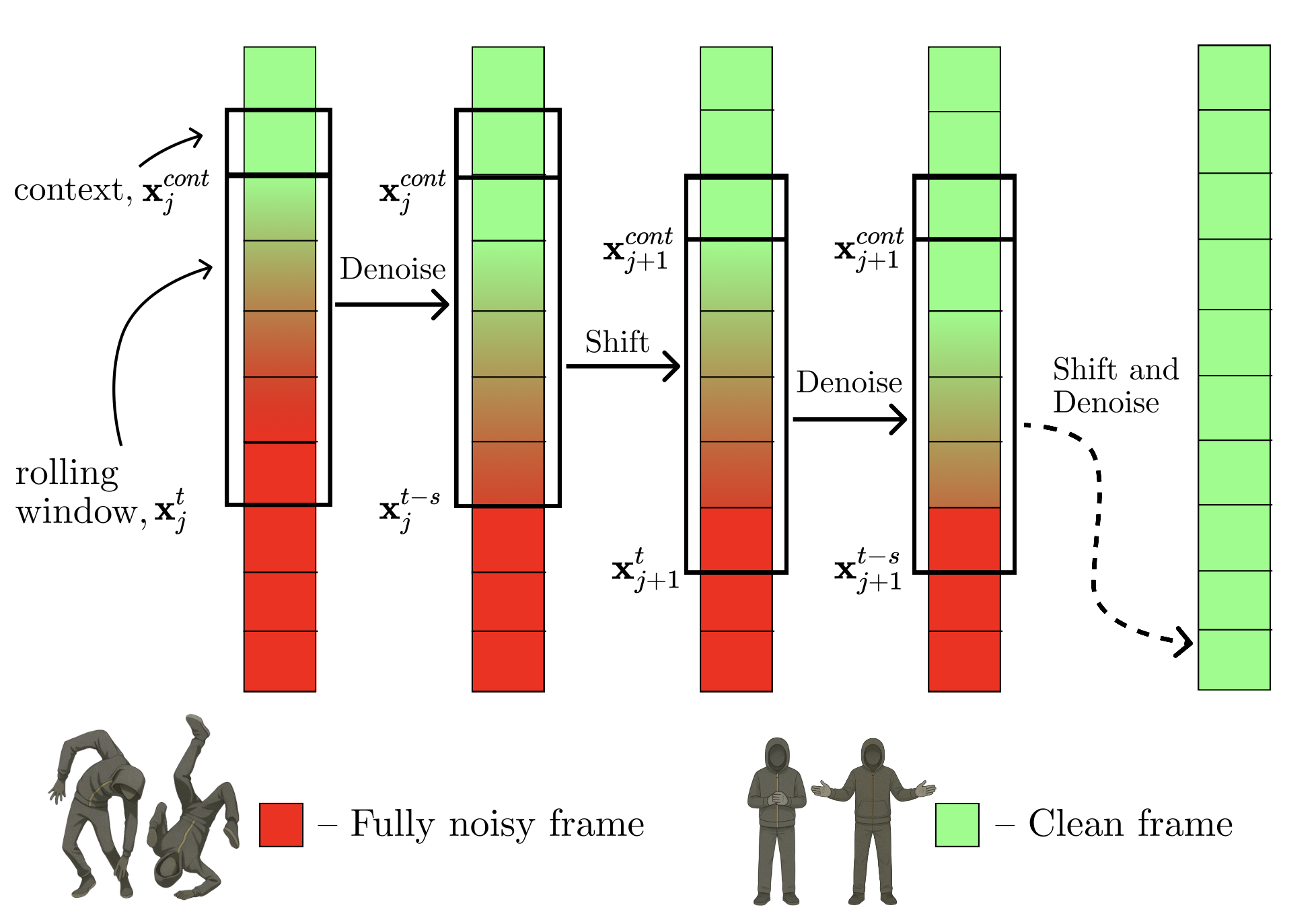}
    \caption{Visualization of the rolling denoising process with parameters $T = 5$, $N = 5$, $n^{cont} = 1$, $s=1$}.
    \label{fig:denoising}
\end{figure} 

Once the first frame is fully denoised, i.e. $t_0$ reaches $0$, a new frame, sampled from Gaussian noise, is introduced, and the window shifts forward ($j=j+1$, $t_0=s$) to continue the denoising process. This approach allows for continuous and unbounded generation, making RDMs particularly effective for producing arbitrarily long sequences. \\

The training objective is \\
\begin{equation}\label{eq:rdm_loss}
 L(\theta) :=\mathbb{E}_j\mathbb{E}_{\mathbf{x}^0_j}\sum_{t^0=1}^s\mathbb{E}_{\mathbf{x}_j^{t_0}} \sum_{n=0}^{N - 1} a(t_n) \| x^0_{j+n} - \hat{x}_{j+n} \|^2, 
\end{equation}
where \( x_k^t \) represents a single frame with the upper index \( t \) denoting the noise level and the lower index \( k \) representing the index in the sequence and $a(t_n)$ is a weighting function (e.g. signal-to-noise ratio). 

To achieve progressive noise scheduling, RDMs operate in two distinct phases: initialization and rolling. In the initialization phase, the model starts with a fully noisy sequence and gradually denoises it to the partially clean state required by the rolling window. Once this point is established, the model enters the rolling phase, where the rolling denoising process described above is applied.

\subsection{Method}\label{sec:method}

In our work, we adapt rolling diffusion models for co-speech gesture generation, introducing a novel framework that transforms any diffusion-based architecture into a streaming model. Our approach enables seamless and continuous gesture generation of arbitrary length by modifying the model architecture and integrating a structured noise scheduling mechanism, which, combined with the rolling denoising process, ensures smooth temporal transitions and prevents abrupt motion discontinuities. As illustrated in figure~\ref{fig:denoising}, the model generates a new clean frame in each $s$-step and shifts the generation window forward to include the new frame at the end.
% \begin{figure}[htb]
%     \centering
%     \includegraphics[width=0.45\textwidth]{images/visualization.pdf}
%     \caption{Visualization of the rolling denoising process with parameters $T = 5$, $N = 5$, $n^{cont} = 1$, $s=1$}
%     \label{fig:denoising}
% \end{figure} 

% In our implementation, we discretize time using \( t \in [0, T] \) instead of the continuous range \( t \in [0,1] \), where \( T = 1000 \) represents the total number of noise steps. Since the generation window size \( N \) is typically much smaller than \( T \), the noise level difference between adjacent frames is greater than 1.  We define this difference as a step \( s \), calculated as
% \(s = \frac{T}{N}\). 
% To ensure uniform noise distribution across frames, we select the generation window as a divisor of \( T \), ensuring consistent step sizes and preventing uneven noise application. This structured noise scheduling allows for a more controlled and stable generation process, improving the overall quality of generated sequences.
As a condition, the model receives audio features as input $U = \{u_k\}_{k = 0}^{L - 1}$. Audio processing depends on the implementation of the baseline model, but the most popular approach is to use a pre-trained model such as WavLM~\cite{chen2022wavlm}. Also, depending on the baseline's architecture, the model can receive a speech style or speaker ID as an additional condition.

The architecture of the baseline model is mostly preserved, and the main change affects the integration of time embedding. All models incorporate time embeddings into each frame, either by summing or concatenating them with the frame embeddings. In the original models, a single time embedding is replicated across all frames. In contrast, our approach assigns a unique time embedding to each frame within the rolling window.

\subsection{Training Description}
During training, we sample the initial noise level \( t_0 \) for the first frame from a uniform distribution \(
t_0 \sim \text{Uniform}(\{1, \dots, s\}) \) and select the starting index \( j \) for the rolling window uniformly \(j \sim \text{Uniform}(\{1, \dots, L - N\})\). We then determine the noise levels for subsequent frames according to
$t_n = t_0 + ns$. These noise levels are applied using \cref{eq:rdm_q}. As a result, the noise level for each frame \( n \) in the sequence falls within the range \(
t_n \in [s \cdot n,\; s \cdot (n + 1)) \).
To improve the performance of the model, we include several clean frames at the beginning of the sequence as an additional context, represented by $\mathbf{x}_j^{cont} = (x_{j - n^{cont}}^1, \dots, x_{j-1}^1)$ with a length $n^{cont}$. We discovered that applying a minimal level of noise \(t=1\) to these context frames is essential. This process acts as a form of regularization and improves stability
(see the Supplementary Material for more details). Therefore, at each time step, the input to the model consists of a concatenated sequence \([\mathbf{x}^{cont}_j, \mathbf{x}^{t}_j]\). Here, \(\mathbf{x}^{t}_j\) denotes a rolling window whose first frame corresponds to the \(t\)-th noise level, and the associated audio features for this sequence are given by \( u_j = (u_{j - n^{cont}}, \dots, u_{j + N - 1}) \). \\
Unlike prior work, our model is trained exclusively for the rolling phase, omitting an initial descent phase. Furthermore, we use \( a(t_n) = 1 \) in the training objective for all \( t \), instead of using the signal-to-noise ratio (SNR). The training procedure is summarized in \cref{training alg}.

\begin{algorithm}[ht]
\caption{Training}
\label{training alg}
\begin{algorithmic}[1]
\Repeat
    \State $j \sim \text{Uniform}(\{1, \dots, L - N\})$
    \State $t_0 \sim \text{Uniform}(\{1, \dots, s\})$
    \State $\mathbf{x}^0_j \gets (x^0_j, \dots, x^0_{j + N - 1})$
    \State $\mathbf{u}_j \gets (u_{j-n^{cont}}, \dots, u_{j + N - 1})$
    \State Sample $\mathbf{x}^{t_0}_j \sim q(\mathbf{x}^{t_0}_j|\mathbf{x}_j)$ (see \cref{eq:rdm_q})
    \State Compute $\hat{\mathbf{x}}_j \gets f_{\theta}\Bigl([\mathbf{x}^{\mathrm{cont}}_j,\, \mathbf{x}^{t_0}_j],\, t_0,\, \mathbf{u}_j\Bigr)$
    \State Take gradient descent step on
    \[
    \nabla_\theta \sum_{n=0}^{N - 1} a(t_n) \| x^0_{j+n} - \hat{x}_{j+n} \|^2
    \]
\Until{Converged}
\end{algorithmic}
\end{algorithm}

\subsection{Sampling Process Description}\label{sec:sampling}
To create a progressively noisy sequence within the rolling window, we begin by padding the sequence with idle poses, each characteristic of the style and corresponding to silence. These initial poses are then noised according to a schedule starting at \(t_0 = s\). At each \(s\)-th step, we get a fully denoised frame, which is appended to the output sequence. Subsequently, a new frame is sampled from Gaussian noise, added to the end of the rolling window, and the window is shifted. The sampling process is outlined in \cref{sampling_alg}.

\begin{algorithm}[ht]
\caption{Sampling}
\label{sampling_alg}
\begin{algorithmic}[1]
\Require audio $\{u_j\}_{j = 0}^{L - 1}$, idle $x_{\mathrm{idle}}$
\Ensure Resulted prediction $y$
\For{$n = -N, \dots, -1$}
    \State $t_n = s (N + n + 1)$ 
    \State $u_n = 0$
    \State Sample $x_n^{t_n} \sim q(x_n^{t_n} | x_{\mathrm{idle}})$
\EndFor
\State $\mathbf{u}_{-N} \gets (u_{-N-n^{cont}}, \dots, u_{-1})$
\State $\mathbf{x}_{-N}^s \gets (x_{-N}^s, \dots, x_{-1}^T)$
\State $\mathbf{x}_{-N}^{\mathrm{cont}} \gets ({x}_{\mathrm{idle}}, \dots, x_{\mathrm{idle}})$
\State $j = -N$
\Repeat
    \For{$t_0 = s, s-1, ..., 1$}
        \State $\hat{\mathbf{x}}_j \gets f_{\theta}\Bigl([\mathbf{x}^{\mathrm{cont}}_j,\, \mathbf{x}^{t_0}_j],\, t_0,\, \mathbf{u}_j\Bigr)$
        \State Sample $\mathbf{x}^{t_0 - 1}_j \sim p_\theta(\mathbf{x}^{t_0 - 1}_l|\mathbf{x}^{t_0}_j)$ (see \cref{eq:rdm_p})
    \EndFor
    \State $y_j = \hat{x}_j^0; j = j + 1$
    \State Sample $x_{j - 1}^1 \sim q(x_{j - 1}^1 | x_{j - 1}^0); x^T_{j+N} \sim \mathcal{N}(0, I)$
    \State $\mathbf{x}_j^{cont} \gets (x^1_{j - n^{\mathrm{cont}}}, \dots, x_{j - 1}^1)$
    \State $\mathbf{x}_j^s \gets (x_{j}^s, \dots, x_{j + N}^T)$
    \State $\mathbf{u}_{j} \gets (u_{j  - n^{cont} }, \dots, u_{j + N})$
\Until{Completed}
\end{algorithmic}
\end{algorithm}

\section{Rolling Diffusion Ladder Acceleration}

\begin{figure}[htb]\
    \centering
    \includegraphics[width=0.47\textwidth]{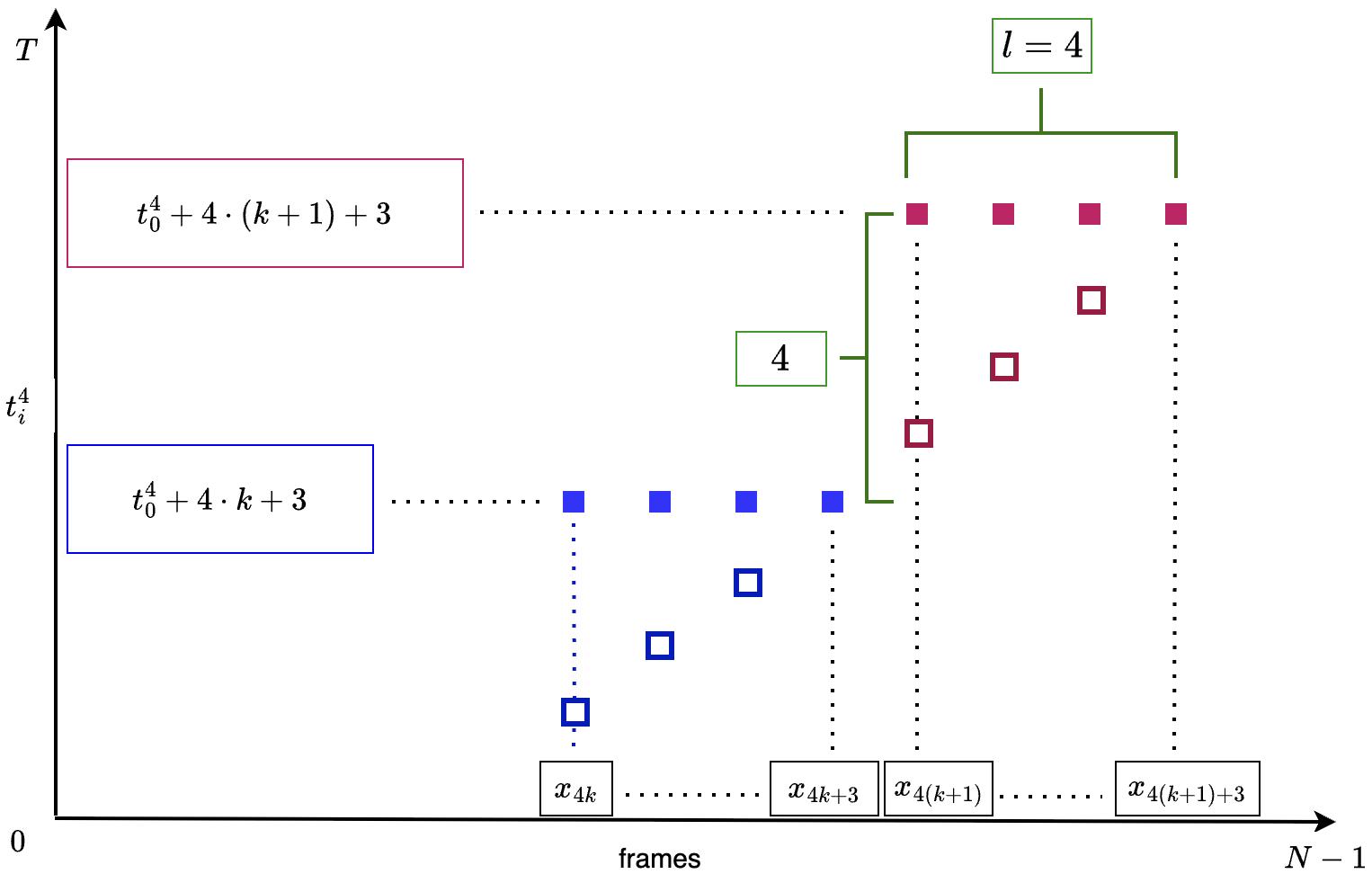}
    \caption{Rolling ladder steps $k$ (blue bottom squares) and $k+1$ (red upper squares) for the ladder step size $l=4$ with corresponding noise level values and frames in the rolling window $\mathbf{x}^{t_0^4}_j$. The hollow squares of the corresponding color show the initial positions of the noise levels for a ladder of step size $l=1$.}
    \label{fig:ladder}
\end{figure}

In the standard rolling diffusion sampling process (Section~\ref{sec:sampling}), only a single frame is fully denoised at each $s$-th step, leading to a sequential bottleneck that slows down the overall generation. Hence we cannot make $T$ smaller than $N$, which corresponds to $s=1$.  To overcome this limitation, we introduce Rolling Diffusion Ladder Acceleration (RDLA), a novel approach that transforms the original noise schedule into a ladder with step size $l$ (see \cref{fig:ladder}), enabling the simultaneous denoising of $l$ frames from the same noise level. We introduce a new sequence of stepsize noise levels $t_i^l$ according to the following rule:
% $$
% t^l_i=t^l_0,\ \ \ kl\le i<(k+1)l-1,\ \ k\in\{0,\frac{N}{l}-1\},
% $$
$$
t^l_i=t_0^l + (k + 1) \cdot l - 1,\ \ \ kl\le i<(k+1)l-1,
$$
where $k\in\{0,\frac{N}{l}-1\}$, $t_0^l\in\{1,\ldots,l\}$.
% $t_0^l \sim \text{Uniform}(\{1, ..., s\}).$

This modification allows multiple frames to be jointly denoised in each iteration, accelerating the process. The conventional rolling diffusion model can be seen as a special case of RDLA with $l=1$. The process of constructing the ladder noise schedule $l=4$ for steps $k$ and $k+1$ is illustrated in \cref{fig:ladder}. This proposed process can be viewed as a transformation from the noise schedule with $l=1$ to the noise schedule with $l > 1$ ($l=4$ for ~\cref{fig:ladder}). All noise levels within a constructing ladder step are set equal to the last noise level in that step. This design choice ensures a consistent step height across all levels of the ladder and guarantees that the last ladder step noise level equals to $T$. Thus, the sampling process in a rolling window starts from $T$, and has zero signal-to-noise ratio, which, as demonstrated in~\cite{lin2024common}, is important for maintaining output quality.  

During inference, RDLA processes $l$ frames simultaneously by fully denoising an entire block at each iteration. At the same time, $l$ new frames are initialized from Gaussian noise and appended to the rolling window, ensuring continuous sequence expansion. This modification follows \cref{sampling_alg}, but with enhanced efficiency due to the structured noise scheduling of RDLA. 

\subsection{RDLA Training Strategy}\label{sec:rdla_train}

To further enhance RDLA’s effectiveness, we introduce a progressive fine-tuning approach where the ladder step size  is gradually increased during training (e.g., $l=2, 4, ...$). The model architecture $f_{\theta}$ remains unchanged, but its weights are initialized from the previous iteration as $l$ increases.

One key challenge is the loss of coherence between context and newly generated frames when $l$ becomes large. To address this, we increase the context window size $N+ n^{\text{cont}}$ while keeping the total rolling window length $N$ fixed. To preserve the divisibility of $T$ by $N$, we also needed to reduce $T$. This ensures stable training while maintaining performance across various step sizes.

Additionally, we introduce an inertial loss function $L_{\text{RDLA}}(\theta)$ to regularize the transition between ladder steps:
% \begin{equation}
%     \begin{split}
%         L_{\text{inert}\;\theta}^{l}(\mathbf{x}, t) = \sum_{k=0}^{N/l} \Bigr[ \sum_{i=0}^{l-1} \| x^0_{lk+i} - \hat{x}_{lk+i}\|^2 - \\
%         - 2\lambda\sum_{i=0}^{l-2}(x^0_{lk+i} - \hat{x}_{lk+i})(x^0_{lk+i+1} - \hat{x}_{lk+i+1}) \Bigl]
%     \end{split}
% \end{equation}
% \begin{multline}\label{eq:rdla_loss}
% \mathbb{E}_j \mathbb{E}_{\mathbf{x}^0_j}  \sum_{t_0^l=1}^s \mathbb{E}_{\mathbf{x}_j^{t_0^l}} \Biggl[ \sum_{k=0}^{N/l - 1} \Biggl( \sum_{i=0}^{l-1} \| x^0_{j+lk+i} - \hat{x}_{j+lk+i} \|^2 \\
% - 2\lambda \sum_{i=0}^{l-2} \langle x^0_{j+lk+i} - \hat{x}_{j+lk+i}, x^0_{j+lk+i+1} - \hat{x}_{j+lk+i+1} \rangle \Biggr) \Biggr]
% \end{multline}
\begin{multline}\label{eq:rdla_loss}
\mathbb{E}_j \mathbb{E}_{\mathbf{x}^0_j}  \sum_{t_0^l=1}^l \mathbb{E}_{\mathbf{x}_j^{t_0^l}} \Biggl[ \sum_{n=0}^{N - 1}  \| x^0_{j+n} - \hat{x}_{j+n} \|^2 \\
- 2\lambda \sum_{n=0}^{N-1} \langle x^0_{j+n} - \hat{x}_{j+n}, x^0_{j+n+1} - \hat{x}_{j+n+1} \rangle \Biggr]
\end{multline}
where the second term penalizes abrupt changes between adjacent denoised frames, reducing jitter.

% During our experiments with RDLA fine-tuning we found that an important cause of tremor is the weakening of the connection between context and frames in a completely denoised step. To compensate for this effect, we increased the context size $n^{\text{context}}$ during RDLA training while maintaining the overall rolling window length $N$. To preserve the divisibility of $T$ by $N - n^{\text{context}}$, we also needed to reduce $T$.

% To improve the results of naive sampling we propose the RDLA fine-tuning procedure in which ladder step size increases iteratively: 2, 4, etc. During this procedure, the model architecture $f_{\theta}$ remains unchanged. At each iteration, as the step size increases, the model undergoes further training, with its weights initialized from the previous iteration. The noise schedule $t^l$ is adjusted to match the current step size $l$. The target for training is the ground truth  $\mathbf{x}$.

% For RDLA training we propose the new inertial loss function for the ladder step size $l$:

% % \begin{equation}
% %     L_{\theta}^{l} = \sum_k \Bigr[ \sum_{i=0}^{l-1} \| x_{lk+i} - \hat{x}_{lk+i}\|^2 - 2\lambda\sum_{i=0}^{l-2}(x_{lk+i} - \hat{x}_{lk+i})(x_{lk+i+1} - \hat{x}_{lk+i+1}) \Bigl]
% % \end{equation}

% \begin{equation}
%     \begin{split}
%         L_{\theta}^{l}(\mathbf{x}, t) = \sum_k \Bigr[ \sum_{i=0}^{l-1} \| x_{lk+i} - \hat{x}_{lk+i}\|^2 - \\
%         - 2\lambda\sum_{i=0}^{l-2}(x_{lk+i} - \hat{x}_{lk+i})(x_{lk+i+1} - \hat{x}_{lk+i+1}) \Bigl]
%     \end{split}
% \end{equation}

By implementing RDLA, we achieve a substantial reduction in inference time while maintaining high visual fidelity and temporal consistency. Empirical results (Section~\ref{sec:results}) demonstrate that RDLA accelerates gesture synthesis by up to 4× compared to standard rolling diffusion.

\begin{figure*}[!t]
  \centering
  \includegraphics[width=\textwidth]{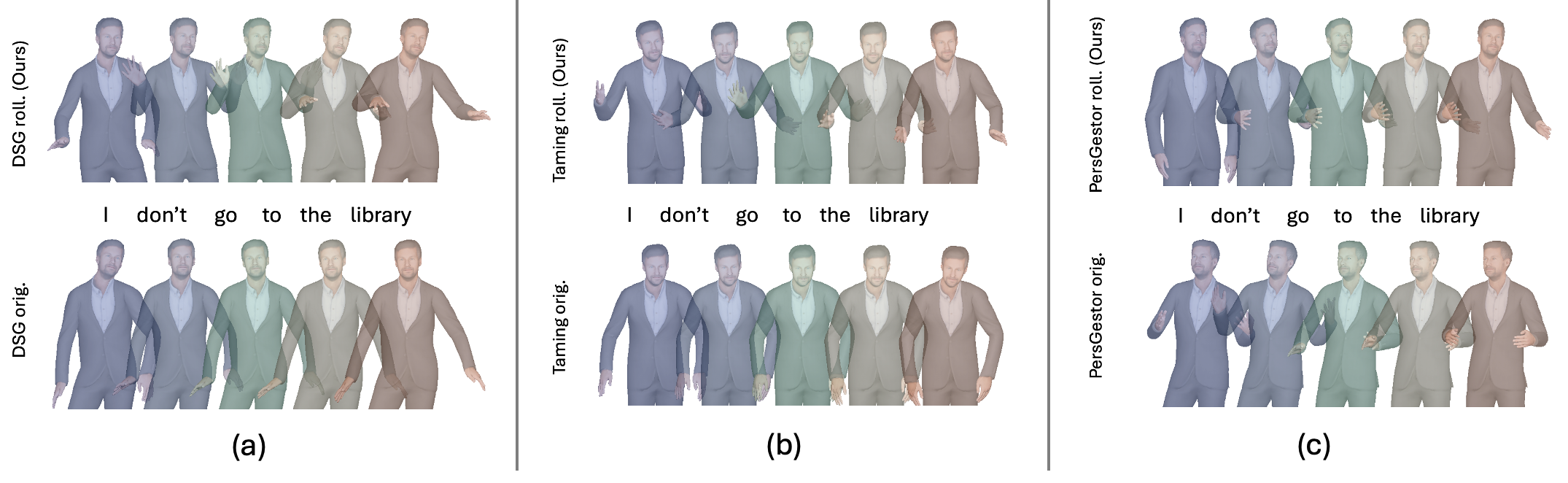}
  \caption{\textbf{Qualitative Comparison on ZEGGS Dataset.} Columns (a–c) are DiffuseStyleGesture, Taming, and PersonaGestor; the baseline is on the bottom row and our version on the top. Our approach generates a broader range of natural, diverse gestures that are tightly synchronized with the speech signal. Specifically, it responds to the negation “don’t” with a rejecting/withdrawing motion, whereas the baselines largely repeat neutral poses and miss this semantic cue.}
  \label{fig:gestures_vis}
\end{figure*}

\section{Experiments}

To thoroughly examine the impact of our method, we integrate our progressive noise scheduling technique into multiple baseline models and conduct comparisons across two datasets: ZEGGS ~\cite{ghorbani2023zeroeggs} and BEAT \cite{liu2022beatlargescalesemanticemotional}. ZEGGS was chosen for its clean, high-quality motion capture data, recorded with a motion capture suit, ensuring precise gesture representation. Additionally, it includes diverse speaking styles, making it well-suited for evaluating stylistic consistency. In contrast, BEAT is one of the largest and most widely used datasets in the field, providing a broader range of conversational gestures. However, it contains more noise, presenting a greater challenge for generative models. Gestures in these datasets are represented in the BioVision Hierarchy (BVH) format and processed as vectors encoding joint positions and rotations in 3D space, along with additional features such as velocities. The processing method used in our work follows the baseline approaches.

% This allows for a systematic evaluation of improvements in gesture generation quality across different architectures. Our goal is to determine whether our approach can serve as a generalizable framework for enhancing diverse gesture synthesis models, ensuring robustness and adaptability across varying data conditions. 

This enables systematic evaluation of gesture-generation improvements across architectures. Our aim is to assess whether the method can serve as a generalizable framework for enhancing diverse gesture-synthesis models, ensuring robustness across varying data conditions. All evaluations used long sequences (1.5–2-min clips from ZEGGS/BEAT), from which metrics and user-study samples were derived.

\subsection{Experimental Setup}

% \textbf{Baseline approaches.}
\subsubsection{Baseline approaches.}
% Taming functions as a general-purpose co-speech gesture model, while DiffuseStyleGesture and DiffSHEG incorporate stylistic control through conditioning on predefined style labels. In contrast, PersonaGestor also accounts for stylistic variations, but derives them from features extracted directly from the audio rather than relying on explicit style labels.
As the primary baselines for our work, we selected state-of-the-art diffusion-based models for gesture generation: Taming Diffusion \cite{zhu2023tamingdiffusionmodelsaudiodriven}, DiffuseStyleGesture \cite{yang2023diffusestylegesture} (DSG), Per\-so\-naGestor \cite{zhang2024speechdrivenpersonalizedgesturesynthetics} and DiffSHEG \cite{chen2024diffsheg}. These models were chosen because they represent leading approaches in data-driven gesture synthesis, each tackling different aspects of the task. Taming functions as a general-purpose co-speech gesture model, while DiffuseStyleGesture and DiffSHEG incorporate stylistic control through conditioning on predefined style labels. In contrast, PersonaGestor also accounts for stylistic variations, but derives them from features extracted directly from the audio rather than relying on explicit style labels. Notably, both DSG and Taming inherently employ direct conditioning on seed poses to ensure smooth transitions during generation. DiffSHEG, on the other hand, uses an outpainting strategy to concatenate adjacent clips, which also enables streaming generation. In each experiment, we adopt the same architecture as the baseline model, incorporating the modifications outlined in \cref{sec:method}. Since the DiffSHEG model was not originally developed with the ZEGGS dataset in mind, and its implementation relies on additional information unavailable in ZEGGS, we adapted and tested the model only on a single dataset where all necessary inputs were provided.

For consistency with prior work, we follow the Dif\-fuse\-Style\-Ges\-ture baseline setup, using six distinct speaking styles from the ZEGGS dataset (Happy, Sad, Neutral, Old, Angry, and Relaxed) across all three baselines. This ensures that our results remain comparable. For the BEAT dataset, we include all 30 available speaking styles, leveraging its extensive stylistic diversity to assess how well our method generalizes across a broader range of conversational gestures. All experiments were conducted on a workstation with an NVIDIA A40 GPU (48 GB), 126 GB RAM, 64-core CPU, and Ubuntu 22.04.

% \textbf{Evaluation metrics}. 
\subsubsection{Evaluation metrics.}
To evaluate the quality of our generated gestures, we utilize the metrics \( \text{FD}_g \), \( \text{FD}_k \), \( \text{Div}_g \), and \( \text{Div}_k \) introduced by Ng et al. \cite{ng2024audiophotorealembodimentsynthesizing}. These metrics are computed in the space of 3D joint coordinates.  The Fréchet distance (FD) metrics, \( \text{FD}_g \) and \( \text{FD}_k \), measure the similarity between the generated gestures and the real motion data. Specifically, \( \text{FD}_g \) evaluates the spatial distribution of poses, while \( \text{FD}_k \) assesses motion dynamics by analyzing frame-to-frame differences. Lower values indicate a closer resemblance to real-world gestures.  The Diversity (Div) metrics, \( \text{Div}_g \) and \( \text{Div}_k \), quantify the variability within the generated gestures. Higher diversity values suggest a richer and more varied set of gestures, preventing repetitive or overly uniform motion. Kinetic-based metrics help ensure that the generated gestures exhibit realistic movement patterns, maintaining a natural motion flow without sudden, unnatural position changes. Together, these evaluation measures provide a comprehensive assessment of both the realism and diversity of the generated co-speech gestures.
% \textbf{Training hyperparameters.}
\subsubsection{Training hyperparameters.}
Across all models, we retain most baseline hyperparameters to ensure fair comparison. However, the generation window length is a critical setting due to framework constraints. For the DSG rolling model, we use $N = 100$, with  $n^{\text{cont}} = 8$ for ZEGGS and 30 for BEAT. Regularization is applied with weight decay $\text{wd} = 0.005$ and dropout 0.2 for ZEGGS, and $\text{wd} = 0.01$ for BEAT. For Taming, we adjust the generation window to $N = 50$, $n^{\text{cont}} = 4$, and increase training to 2000 epochs. In PersonaGestor, we use $N = 200$, $n^{\text{cont}} = 20$, and $\text{wd} = 0.01$ for both datasets. The rolling DiffSHEG model uses $N = 25$ and $n^{\text{cont}} = 9$ to match the baseline window length of 34. We preserve all inference settings and do not use resampling for last timesteps. Training is run for 700 epochs. All other hyperparameters remain unchanged.

\subsubsection{RDLA experiments.}
To evaluate the effectiveness of Rolling Diffusion Ladder Acceleration (RDLA) in improving inference efficiency, we conducted a series of experiments on ZEGGS and BEAT dataset. To ensure a fair comparison, we applied RDLA to the DiffuseStyleGesture (DSG) model~\cite{yang2023diffusestylegesture}, which has demonstrated state-of-the-art performance among diffusion-based methods on these benchmarks.

A straightforward method for accelerating the sampling process is reducing the number of denoising steps per frame. As described in \cref{sec:sampling}, in our framework, each frame undergoes a predefined number of denoising steps $s$ before the rolling window shifts forward in time. In RDLA we reducing this number from $s$ to 1 leading to a total inference step count of $T_r=N$, where $N$ is the total number of frames in sliding window. As a result, we report RDLA results with a reduced number of denoising steps ($T_r=100$) using DDPM~\cite{ho2020denoising} sampling strategy, while additional experiments with varying $T_r$ values and alternative sampling strategies are provided in the Supplementary Material.

% Beyond reducing the number of sampling steps, we leveraged RDLA to accelerate inference in the temporal dimension. By constructing a denoising ladder with step size $l$, we simultaneously denoised $l$ frames at each iteration, achieving an $l$-fold acceleration. Our experiments applied RDLA to DSG on ZEGGS and BEAT datasets with ladder step sizes $l \in \{2, 4\}$. For RDLA fine-tuning we used model's weights obtained in \cref{sec:results} for $l=1$ as initial and continue training for $3000$ epochs with $\text{learning rate}=1e-7$, $\text{dropout}=0.1$, $\text{wd}=0.01$, and all remaining parameters followed the original DSG setup. During our experiments, we observed that RDLA performs best when the context length increases proportionally with the ladder step size. Accordingly, we report results using a context length of $n^{\text{cont}} = 28$ for ZEGGS and $n^{\text{cont}} = 50$ for BEAT. Further details are provided in the Supplementary Material.

Beyond reducing the number of sampling steps, we leveraged RDLA to accelerate inference in the temporal dimension. By constructing a denoising ladder with step size $l$, we simultaneously denoised $l$ frames at each iteration, achieving an $l$-fold acceleration. Our experiments applied RDLA to DSG on ZEGGS and BEAT datasets with ladder step sizes $l \in \{2, 4\}$. For RDLA fine-tuning we used model's weights obtained in \cref{sec:results} for $l=1$ as initial and continue training for $3000$ epochs with $\text{learning rate}=1e-7$, $\text{dropout}=0.1$, $\text{wd}=0.01$, and all remaining parameters followed the original DSG setup. During our experiments, we observed that RDLA performs best when the context length increases proportionally with the ladder step size. Accordingly, we report results using a context length of $n^{\text{cont}} = 28$ for ZEGGS and $n^{\text{cont}} = 50$ for BEAT. Minor tremor that we observed for larger $l$ is mitigated via the temporal-smoothness loss~(\cref{eq:rdla_loss}) and an On-the-Fly Smoothing strategy, though the latter was not used in reported results. Further details are provided in the Supplementary Material.

\begingroup
\setlength{\tabcolsep}{0.25pt}      % default is 6pt
\renewcommand{\arraystretch}{0.2}% tighten rows a bit

\begin{table}[t]
  \centering\footnotesize
  \begin{tabular}{@{}l@{\hspace{2pt}}cccc@{}}
    \toprule
    Method & $\text{Div}_g \uparrow$ & $\text{Div}_k \uparrow$ & $\text{FD}_g \downarrow$ & $\text{FD}_k \downarrow$ \\
    \midrule
    GT              & 272.34          & 213.97          & \multicolumn{1}{c}{--} & \multicolumn{1}{c}{--} \\
    \midrule
    \m{DSG}{orig.}       & $239.37_{\pm2.4}$ & $161.07_{\pm3.3}$ & $6393.99_{\pm39.0}$    & $14.24_{\pm1.2}$ \\
    [0.6ex]
    \hdashline
    \addlinespace[0.6ex]
    \m{DSG}{roll.}       & $\mathbf{251.35}_{\pm 0.8}$ & $\mathbf{175.12}_{\pm 1.8}$          & $\mathbf{3831.35}_{\pm 91.2}$       & $\mathbf{8.08}_{\pm 0.7}$    \\
    [0.6ex]
    \hdashline
    \addlinespace[0.6ex]   
    \m{DSG}{RDLA 2}      & $222.25_{\pm 1.2}$          & $173.76_{\pm 1.0}$          & $5772.40_{\pm 63.9}$                & $13.65_{\pm 0.1}$ \\
    [0.6ex]
    \hdashline
    \addlinespace[0.6ex]   
    \m{DSG}{RDLA 4}     & $157.77_{\pm 0.5}$          & $93.22_{\pm 0.6}$           & $16791.03_{\pm 59.4}$               & $42.19_{\pm 1.5}$            \\
    \midrule
    \m{Taming}{orig.}    & $154.70_{\pm 1.2}$          & $80.70_{\pm 0.8}$           & $10784.86_{\pm 54.7}$               & $418.85_{\pm 12.7}$           \\
    [0.6ex]
    \hdashline
    \addlinespace[0.6ex]   
    \m{Taming}{roll.}    & $\mathbf{190.09}_{\pm 1.2}$          & $\mathbf{124.42}_{\pm 1.0}$          & $\mathbf{9064.00}_{\pm 102.7}$                & $\mathbf{353.62}_{\pm 23.5}$           \\
    \midrule
    \m{Pers\-Gestor}{orig.}& $230.11_{\pm 6.0}$          & $165.17_{\pm 1.9}$          & $4060.36_{\pm 68.2}$                & $11.12_{\pm 1.6}$            \\
    [0.6ex]
    \hdashline
    \addlinespace[0.6ex] 
    \m{Pers\-Gestor}{roll.}& $\mathbf{242.14}_{\pm 7.8}$          & $\mathbf{189.31}_{\pm 2.7}$ & $\mathbf{3936.75}_{\pm 56.2}$                & $\mathbf{9.14}_{\pm 0.5}$             \\
    \bottomrule
  \end{tabular}
  \caption{Results of quantitative analysis on ZEGGS dataset.}
  \label{tab:ZEGGS_res}
\end{table}
\endgroup

\begingroup
\setlength{\tabcolsep}{0.25pt}      % default is 6pt
\renewcommand{\arraystretch}{1.2}   % tighten rows a bit

\begin{table}[t]
  \centering\footnotesize
  \begin{tabular}{@{}l@{\hspace{2pt}}cccc@{}}
    \toprule
    Method & $\text{Div}_g \uparrow$ & $\text{Div}_k \uparrow$ & $\text{FD}_g \downarrow$ & $\text{FD}_k \downarrow$ \\
    \midrule
    GT               & 279.52          & 116.17          & \multicolumn{1}{c}{--} & \multicolumn{1}{c}{--} \\
    \midrule
    \m{DSG}{orig.}        & $201.83_{\pm 8.4}$        & $63.78_{\pm 5.7}$         & $37062.19_{\pm 459.9}$       & $77.28_{\pm 1.9}$ \\
    [0.6ex]
    \hdashline
    \addlinespace[0.6ex]   
    \m{DSG}{roll.}        & $\mathbf{241.50}_{\pm 9.8}$ & $76.09_{\pm 6.1}$         & $21441.91_{\pm 434.0}$       & $69.23_{\pm 5.7}$ \\
    [0.6ex]
    \hdashline
    \addlinespace[0.6ex]   
    \m{DSG}{RDLA 2}       & $190.41_{\pm 9.9}$        & $58.72_{\pm 8.4}$ & $\mathbf{17309.63}_{\pm 327.1}$       & $\mathbf{56.24}_{\pm 5.5}$ \\
    [0.6ex]
    \hdashline
    \addlinespace[0.6ex]   
    \m{DSG}{RDLA 4}       & $207.47_{\pm 9.1}$        & $\mathbf{83.08}_{\pm 8.7}$ & $18739.77_{\pm 467.2}$       & $59.02_{\pm 4.9}$ \\
    \midrule
    \m{Taming}{orig.}      & $139.64_{\pm 0.6}$        & $57.40_{\pm 0.5}$         & $11632.64_{\pm 158.7}$       & $\mathbf{67.94}_{\pm 3.7}$ \\
    [0.6ex]
    \hdashline
    \addlinespace[0.6ex]   
    \m{Taming}{roll.}     & $\mathbf{169.59}_{\pm 1.6}$        & $\mathbf{73.41}_{\pm 1.4}$         & $\mathbf{9835.27}_{\pm 190.7}$ & $74.15_{\pm 9.2}$ \\
    \midrule
    \m{PersGestor}{orig.} & $173.05_{\pm 3.7}$        & $51.54_{\pm 1.6}$         & $11936.82_{\pm 324.5}$       & $143.37_{\pm 5.1}$ \\
    [0.6ex]
    \hdashline
    \addlinespace[0.6ex]   
    \m{PersGestor}{roll.} & $\mathbf{181.12}_{\pm 3.2}$        & $\mathbf{62.50}_{\pm 1.5}$         & $\mathbf{10815.76}_{\pm 350.4}$       & $\mathbf{130.14}_{\pm 5.2}$ \\
    \midrule
    \m{DiffSHEG}{orig.} & $247.47_{\pm 3.0}$        & $108.7_{\pm 2.0}$         & $13547.4_{\pm 835}$       & $4517.2_{\pm 172}$ \\
    [0.6ex]
    \hdashline
    \addlinespace[0.6ex]   
    \m{DiffSHEG}{roll.} & $\mathbf{294.77}_{\pm 4.2}$        & $\mathbf{110.8}_{\pm 2.1}$         & $\mathbf{10201.9}_{\pm 1586}$       & $\mathbf{1696.6}_{\pm 307}$ \\
    \bottomrule
  \end{tabular}
  \caption{Results of quantitative analysis on BEAT dataset.}
  \label{tab:BEAT_res}
\end{table}
\endgroup

\subsection{Results}\label{sec:results}
The quantitative evaluation on ZEGGS and BEAT datasets is summarized in \cref{tab:ZEGGS_res} and \cref{tab:BEAT_res} (values are reported as mean$_{\pm\text{std}}$, where the mean and standard deviation are computed over five independent runs).
Across both datasets, rolling variants generally outperform their original counterparts, with statistically significant gains in Fréchet distance and diversity (paired $t$-test, $p<0.05$), except $\text{FD}_k$ for Taming. DSG rolling shows the most notable improvement, while other rolling versions also exhibits enhanced performance. Since our method surpasses baselines that utilize direct conditioning on seed frames or outpainting strategy, it clearly demonstrates the superiority of our approach for generating coherent and high-quality motion. These results confirm that our rolling diffusion framework not only improves generation quality but does so in a model-agnostic manner. By applying our method across multiple state-of-the-art baselines without modifying their core architectures, we show that the framework generalizes well and can serve as a plug-and-play solution for streaming gesture synthesis. 

% Results for RDLA in \cref{tab:ZEGGS_res} demonstrate that a 2× acceleration ($l=2$) yields only a modest drop in performance, maintaining metrics comparable to the baseline on ZEGGS. On the BEAT dataset, as shown in \cref{tab:BEAT_res}, RDLA achieves improved Fréchet distances for both $l=2$ and $l=4$, with a further gain in kinetic diversity at 4× acceleration. These results suggest that RDLA enables significant speedups while preserving or even enhancing perceptual and motion quality.

Results in \cref{tab:ZEGGS_res} show that a 2× acceleration ($l=2$) yields only a modest performance drop on ZEGGS, with metrics close to the baseline. On BEAT (\cref{tab:BEAT_res}), RDLA improves Fréchet distances for both $l=2$ and $l=4$, and further boosts kinetic diversity at 4× acceleration. Thus, RDLA provides substantial speedups while preserving or even improving perceptual and motion quality. The differing trends stem from dataset characteristics: ZEGGS is more expressive, whereas BEAT is calmer; increasing $l$ introduces smoothing that is more disruptive for ZEGGS.

\subsubsection{Acceleration findings.}
% Our rolling modification of DSG achieves a generation speed of 10 FPS on a single GPU from the workstation described above. With a ladder step of $l=1$ and $T_r=100$, the speed increases to 70 FPS. This configuration reduces latency while maintaining the same throughput as the original DSG method, making it well-suited for interaction with streaming audio. Increasing the ladder step to $l=2$ and $l=4$ further boosts performance to 120 FPS and 200 FPS, respectively.

% Our rolling modification of DSG achieves a generation speed of 10 FPS on a single GPU from the workstation described above and matches with the baseline’s throughput. With a ladder step of $l=1$ and $T_r=100$, the speed increases to 70 FPS. This configuration reduces latency while maintaining the same throughput as the original DSG method: it emits the full 4 sec window every 8 sec, while our model streams a new frame every 0.06 sec (0.006 sec at $T_r=100$). Increasing $l=2$ and $l=4$ further boosts performance to 120 FPS and 200 FPS, respectively. Full inference time comparison is in Supplementary Material.

\begin{table}[t]
\centering
\small
\begin{tabular}{lcccc}
\toprule
\textbf{Method} & \textbf{$l$} & \textbf{Steps} & \textbf{FPS} & \textbf{Latency (sec)} \\
\midrule
DSG orig. (baseline) & -- & 1000 & 10 & 8 \\
DSG roll. (ours) & 1 & 1000 & 10 & 0.06 \\
DSG roll. (ours) & 1 & 100 & 70 & 0.006 \\
DSG RDLA (ours) & 2 & 100 & 120 & 0.003 \\
DSG RDLA (ours) & 4 & 100 & 200 & 0.002 \\
\bottomrule
\end{tabular}
\caption{Generation speed comparison.}
\label{tab:speed}
\end{table}

On a NVIDIA A40 (48 GB), our rolling DSG and RDLA variants greatly reduce latency without sacrificing throughput, as summarized in table~\ref{tab:speed}.

\subsubsection{Qualitative comparisons.} Figure~\ref{fig:gestures_vis} presents representative keyframes comparing our method with baseline versions of DSG, Taming, and PersonaGestor. As illustrated, our approach produces more expressive and semantically aligned gestures, such as clear withdrawal motions in response to negations. Full video visualizations are provided in the Supplementary Material for a comprehensive qualitative comparison.

\subsection{User Study}\label{sec:user_study}

To assess the quality of our generated co-speech gestures, we conducted a user study using pairwise comparisons between our model and a baseline. We selected the ZEGGS dataset for its clear and expressive gestures, which allow for a precise evaluation of movement quality, stylistic consistency, and synchronization with speech. We used the DSG model as a baseline for comparison.  

Participants were shown pairs of 15-second videos, each synchronized with the same audio but generated using different models. Both videos were displayed simultaneously, with their positions randomized in each trial to eliminate possible bias towards one side. The participants were asked to compare the two animations and rate them based on style consistency, naturalness and fluidity of animations, audio-animation synchronization, presence of technical issues (such as the gluing effect). The rating could take values $\{-2, -1, 0, 1, 2\}$ where $-2$ indicates a strong preference for the baseline, $-1$ -- slight baseline preference, $0$ means no noticeable differences, $1$ indicates a slight preference for our model and $2$ indicates a strong preference for our model. To ensure the reliability of our evaluators, we included some videos multiple times and verified that the assessors provided consistent ratings before including their responses in the final analysis. The study was carried out on 60 pairs of video (10 per style in six different styles). Twenty-two professional assessors trained to work with video data were recruited to participate.

% \begin{figure}[htb]\
%     \centering
%     \includegraphics[width=0.47\textwidth]{images/user_study_main_1.png}
%     \caption{User study results. ``Ours" means DSG rolling modification, ``Theirs" means original DSG. In total $48.4\%$ of participants preferred our model while $36.3\%$ preferred original DSG.}
%     \label{fig:user_study_main_1}
% \end{figure}
\begin{figure}[htb]\
    \centering
    \includegraphics[width=0.475\textwidth]{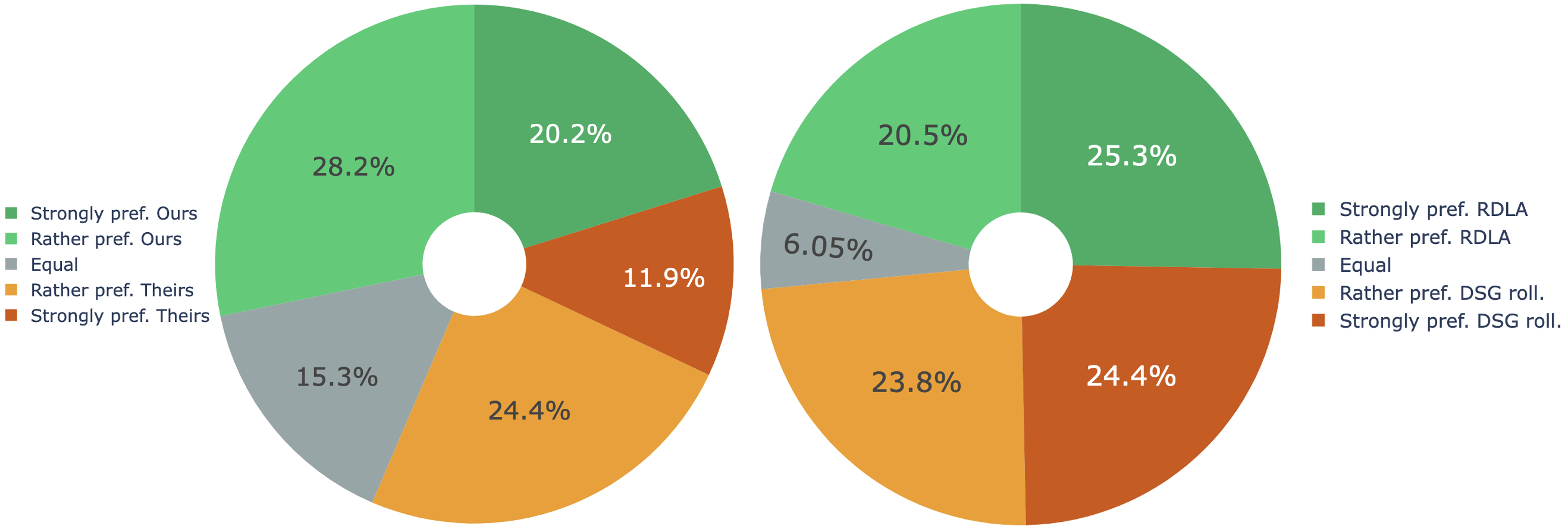}
    \caption{User study results. \textbf{Left:} ``Ours" means DSG rolling modification, ``Theirs" means original DSG. In total $48.4\%$ of participants preferred our model while $36.3\%$ preferred original DSG. \textbf{Right:} RDLA user study results. In total $48.2\%$ of participants preferred DSG rolling model while $45.7\%$ preferred RDLA.}
    \label{fig:user_study_main_1}
\end{figure}

The distribution of the user study results is shown in \cref{fig:user_study_main_1} (Left). Our rolling modification of DSG significantly outperforms the original DSG, aligning well with the quantitative evaluation results; the preference is statistically significant (two-sided Wilcoxon signed-rank, $p<0.05$). To compare RDLA with our original method, we conducted a user study between DSG rolling and RDLA with $l=2$. The distribution of the RDLA user study results is shown in \cref{fig:user_study_main_1} (Right). The RDLA approach is only slightly inferior to our original method, which is consistent with the quantitative findings.

\section{Ablation Study}

In our ablation study (see details in the Supplementary Material), we analyzed key components influencing the Accelerated Rolling Diffusion framework’s performance. First, we evaluated the necessity of minimal noise in context frames. Results showed that incorporating minimal noise ($\sigma^2_1=0.00004$) into context frames significantly improved model robustness and prevented overfitting, thus enhancing generalization. Next, we assessed loss weighting strategies, comparing clamped-SNR~\cite{barrault2024large} weighting versus uniform weighting. Uniform weighting ($a(t) = 1$ in \cref{eq:rdm_loss}) provided a simpler, more stable training process without compromising performance, confirming its suitability for efficient sequential generation. Additionally, we conducted ablations on the components of RDLA: inertial loss regularization~(\cref{eq:rdla_loss}), progressive fine-tuning, on-the-fly smoothing (OFS, not used in the reported results). Removing $L_{\text{RDLA}}(\theta)$ increased motion jitter, underscoring its importance for coherent motion. Progressive fine-tuning was crucial, as models without it exhibited significant quality and diversity loss. Omitting OFS only slightly degraded gesture smoothness. This strategy enabled effective adaptation to increased ladder step sizes, preserving performance.

\section{Conclusion}

We introduced Accelerated Rolling Diffusion, a framework for real-time, high-quality co-speech gesture generation. Experiments on ZEGGS and BEAT, plus user studies, demonstrate strong generalizability and consistent gains across diffusion-based gesture models. Our method produces continuous, temporally coherent motion without post-processing. To boost efficiency, we proposed Rolling Diffusion Ladder Acceleration (RDLA), which uses structured noise scheduling to reduce sampling steps and achieve up to 4× speedup while maintaining realism. RDLA is a promising direction for efficient diffusion acceleration, with potential extensions to other temporal modalities such as text and video.

% {
%     \small
%     \bibliographystyle{ieeenat_fullname}
%     \bibliography{main}
% }

\newpage

\bibliography{aaai2026}

\newpage

\appendix
\section*{Supplementary Material}

\section{Related Work}
\textbf{Long-sequence Motion Generation}. Generating long gesture sequences is challenging due to variable context-driven lengths and memory constraints. Gestures are typically conditioned by text, music, or speech, each requiring tailored approaches. Text-based methods often generate segments individually, stitching them together using weakly supervised techniques with motion smoothness priors \cite{mao2022weaklysupervisedactiontransitionlearning, NEURIPS2023_2d52879e}, iterative refinement like TEACH \cite{10044438}, or diffusion-based methods (e.g., DoubleTake) for temporal consistency \cite{shafir2023humanmotiondiffusiongenerative}. Precise temporal control is provided by multi-track timelines \cite{petrovich2024multitracktimelinecontroltextdriven} and blended positional encodings \cite{barquero2024seamlesshumanmotioncomposition}. Additionally, hierarchical models like MotionMamba \cite{zhang2024motionmambaefficientlong} and MultiAct \cite{Lee_Moon_Lee_2023}, along with language-driven frameworks such as MotionGPT \cite{NEURIPS2023_3fbf0c1e}, enable seamless, coherent generation across extended sequences. \\
\textbf{Dancing Motions Generation.} Long-sequence dance synthesis poses challenges due to duration and style diversity. Methods like EDGE \cite{tseng2022edgeeditabledancegeneration} align overlapping clips for smooth transitions, while "You Never Stop Dancing" \cite{NEURIPS2022_40bfe617} uses low-dimensional manifolds with RefineBank and TransitBank to maintain fluid, high-quality motion. \\
\textbf{Co-speech Gesture Generation}. Driven by the pursuit of natural and expressive human-computer interactions, co-speech gesture generation has rapidly advanced using deep generative techniques. Researchers have explored methods from GANs, VAEs, and VQ-VAEs to diffusion models enhanced with transformer attention. While GANs \cite{habibie2021learningspeechdriven3dconversational} often face instability and mode collapse, diffusion models deliver robustness, high fidelity, and diverse outputs. For instance, DiffuseStyleGesture \cite{yang2023diffusestylegesture} integrates cross-local and self-attention to synchronize varied gestures, and EMAGE \cite{Liu_2024_CVPR} employs masked audio-conditioned modeling with VQ-VAEs for greater expressivity. TalkSHOW \cite{yi2023generatingholistic3dhuman} separately generates facial, body, and hand motions for nuanced speech alignment, while DiffTED \cite{hogue2024difftedoneshotaudiodrivente} uses TPS keypoints in a diffusion pipeline to improve coherence. Lastly, Audio to Photoreal Embodiment \cite{ng2024audiophotorealembodimentsynthesizing} combines diffusion with vector quantization to create realistic conversational avatars. \\
\textbf{Long-sequence Co-Speech Gestures Generation} Generating coherent long gesture sequences is challenging due to potential discontinuities and temporal inconsistencies. A common approach is to generate short chunks separately and then stitch overlapping clips together seamlessly \cite{10.1145/3592458, Liu_2024_CVPR, yang2023diffusestylegesture, zhu2023tamingdiffusionmodelsaudiodriven, zhang2024speechdrivenpersonalizedgesturesynthetics}. For example, FreeTalker \cite{yang2024freetalkercontrollablespeechtextdriven} employs the DoubleTake blending technique to ensure smooth transitions, while autoregressive methods condition each new gesture on preceding outputs to preserve temporal coherence. Additionally, frameworks like DiffSHEG \cite{chen2024diffsheg} use outpainting strategies to extend sequences incrementally, and DiffTED \cite{hogue2024difftedoneshotaudiodrivente} generates coherent TPS keypoint sequences that bridge gesture synthesis with realistic video rendering.\\
Current methods have made progress in long-sequence generation but still face key limitations. They rely on complex architectures with extra conditioning, resulting in high computational demands. Techniques using overlapping segments tend to regenerate identical frames, reducing efficiency. Moreover, most models focus solely on recent frames, neglecting broader temporal context, and their dependence on fully pre-recorded audio and postprocessing limits their use in real-time applications. \\
\textbf{Diffusion Models} have gained popularity for generative tasks, particularly in image and video synthesis. Early work introduced denoising score matching \cite{song2020generativemodelingestimatinggradients} and later evolved into denoising diffusion probabilistic models (DDPM) \cite{ho2020denoising}. Their success in high-quality image synthesis, seen in models like Imagen \cite{saharia2022photorealistictexttoimagediffusionmodels} and Stable Diffusion \cite{rombach2022highresolutionimagesynthesislatent}, has been extended to video generation \cite{ho2022videodiffusionmodels, bartal2024lumierespacetimediffusionmodel, ho2022imagenvideohighdefinition, blattmann2023stablevideodiffusionscaling}. Given their temporal consistency and ability to model complex distributions, diffusion models are also being explored for co-speech gesture generation. \\
\textbf{Rolling Diffusion Models} \cite{ruhe2024rollingdiffusionmodels} extend traditional diffusion-based generation to sequential data, enabling autoregressive synthesis by iteratively generating and conditioning on previous outputs. This approach enhances temporal consistency and long-range dependencies. Recent works \cite{NEURIPS2024_2aee1c41, kim2024fifodiffusiongeneratinginfinitevideos, guo2025dynamical} continue to refine this concept making diffusion models more effective for sequential data generation by enhancing temporal consistency, enabling infinite-length generation, and improving temporal dynamics modeling. 

\section{Rolling Diffusion Ladder Acceleration}

\subsection{On-the-Fly Smoothing}

Our experiments revealed that although the RDLA approach significantly accelerates sampling, it may introduce motion artifacts in certain cases. Specifically, inconsistencies between denoised frame blocks may lead to tremors in motion, degrading both quantitative metrics and visual quality. To mitigate this, we introduce an On-the-Fly Smoothing (OFS) procedure, which refines transitions between consecutive denoised blocks. The core idea is to smooth the transition between the last frame of the previous block and the first frame of the newly denoised block. For $k \in \{0,1,\ldots,l/2\}$:
$$
  \hat{x}_{n^{\text{cont}}+2k-1}=\begin{cases}
    \hat{x}_{n^{\text{cont}}+2k-1}, & \text{if $d_c < \tau$},\\
    (\hat{x}_{n^{\text{cont}}+2k-2} + \hat{x}_{n^{\text{cont}}+2k}) / 2, & \text{otherwise}.
  \end{cases}
$$
where $d_c=\cos{(\hat{x}_{n^{\text{cont}}+2k-2}, \hat{x}_{n^{\text{cont}}+2k})}$, while $\tau$ is a predefined threshold controlling the degree of smoothing. The underlying intuition is that abrupt motion variations are most pronounced when successive frames exhibit high similarity, leading to a visual trembling effect. By averaging adjacent frames when necessary, OFS effectively suppresses these tremors without sacrificing the temporal integrity of motion sequences.

\section{Rolling Diffusion Ladder Acceleration Results}

\subsection{Rolling Diffusion Ladder Acceleration Results}\label{sec:rdla_results}

To evaluate the effectiveness of Rolling Diffusion Ladder Acceleration (RDLA) in improving inference efficiency, we conducted a series of experiments on the ZEGGS dataset~\cite{ghorbani2023zeroeggs}. This dataset was selected due to its high fidelity, as it is obtained through motion capture, minimizing noise and annotation errors. To ensure a fair comparison, we applied RDLA to the DiffuseStyleGesture (DSG) model~\cite{yang2023diffusestylegesture}, which has demonstrated state-of-the-art performance among diffusion-based methods on this benchmark. Following the evaluation protocol in~\cite{yang2023diffusestylegesture}, we tested the performance of models on the ZEGGS validation set, which includes six distinct styles: Happy, Sad, Neutral, Old, Angry, and Relaxed. This results in a total of 36 audio samples for evaluation. In addition to the metrics presented in the main paper, we also report static and kinetic mean squared error (MSE) metrics here.

Our investigation focused on two key aspects of acceleration: (1) reducing the number of denoising steps to minimize computational overhead while maintaining output quality and (2) temporal acceleration via RDLA, which employs ladder-based noise scheduling to denoise multiple frames in each iteration, significantly enhancing inference speed.

\subsubsection{Accelerated inference via fewer diffusion steps.}
A straightforward method for accelerating the sampling process is reducing the number of denoising steps per frame. In our framework, each frame undergoes a predefined number of denoising steps $s$ before the rolling window shifts forward in time. We experimented with reducing this number $s_r$ from $s$ to 1 leading to a total inference step count of $T_r=s_r \cdot N$, where $N$ is the total number of frames in sliding window.

We compared the performance of our accelerated approach against the original DSG method under the same total denoising steps $T_r$. Additionally, we evaluated both DDPM~\cite{ho2020denoising} and DDIM~\cite{song2021denoising} sampling strategies, as DDIM is known to outperform DDPM when the number of steps gets smaller. Experiments were conducted with $N=100$, $n^{\text{cont}}=8$. 

\begin{table}
  \centering
  \begin{tabular}{@{}lcccc@{}}
    \toprule
    Method, $T_r$ & $\text{Div}_g \uparrow$ & $\text{Div}_k \uparrow$ & $\text{FD}_g \downarrow$ & $\text{FD}_k \downarrow$ \\
    \midrule
    GT & 272.34 & 213.97 & - & -  \\
    \midrule
    $\text{DSG}_{\text{DDPM}}$, $1000$ & 239.37 & 161.07  & 6393.99 & 14.24 \\
    $\text{Ours}_{\text{DDPM}}$, $1000$ & \bf{251.35} & \bf{175.12} & \bf{3831.35} & \bf{8.08} \\
    $\text{Ours}_{\text{DDIM}}$, $1000$ & \bf{244.69} & \bf{67.9} & \bf{3559.6} & \bf{10.6} \\
    \midrule
    $\text{Ours}_{\text{DDPM}}$, $500$ & \bf{253.95} & \bf{182.95} & \bf{3147.35} & \bf{9.82} \\
    $\text{Ours}_{\text{DDIM}}$, $500$ & 245.65 & 168.85 & 3607.79 & 10.40 \\
    \midrule
    $\text{DSG}_{\text{DDPM}}$, $100$  & 236.43 & 160.42 & 5647.32 & 12.01 \\
    $\text{DSG}_{\text{DDIM}}$, $100$  & 231.42 & 151.60 & 6148.46 & 11.96 \\
    $\text{Ours}_{\text{DDPM}}$, $100$ & \bf{256.03} & \bf{179.33} & 3612.14 & 11.74 \\
    $\text{Ours}_{\text{DDIM}}$, $100$ & 244.95 & 168.59 & \bf{3475.69} & \bf{11.51} \\
    \bottomrule
  \end{tabular}
  \caption{Performance evaluation with reduced sampling steps on ZEGGS dataset. For each $T_r$ category the best values highlighted in bold.}
  \label{tab:reduse_t_1}
\end{table}

\begin{table}
  \centering
  \begin{tabular}{lcc}
    \toprule
    Method, $T_r$ & $\text{MSE}_s \downarrow$ & $\text{MSE}_k \downarrow$ \\
    \midrule
    $\text{DSG}_{\text{DDPM}}$, $1000$ & 23.42 & 143.13 \\
    $\text{Ours}_{\text{DDPM}}$, $1000$ & 24.89 & \bf{121.76} \\
    $\text{Ours}_{\text{DDIM}}$, $1000$ & 25.02 & 132.54 \\
    \midrule
    $\text{Ours}_{\text{DDPM}}$, $500$ & 24.48 & 143.96 \\
    $\text{Ours}_{\text{DDIM}}$, $500$ & 24.27 & 137.17 \\
    \midrule
    $\text{DSG}_{\text{DDPM}}$, $100$  & \bf{23.24} & 141.05 \\
    $\text{DSG}_{\text{DDIM}}$, $100$  & \bf{23.08} & 138.26 \\
    $\text{Ours}_{\text{DDPM}}$, $100$ & 24.64 & 137.17 \\
    $\text{Ours}_{\text{DDIM}}$, $100$ & 24.52 & \bf{135.81} \\
    \bottomrule
  \end{tabular}
  \caption{Performance evaluation with reduced sampling steps on ZEGGS dataset using MSE metrics.}
  \label{tab:reduse_t_2}
\end{table}

As shown in \cref{tab:reduse_t_1} and \cref{tab:reduse_t_2}, reducing the number of denoising steps generally maintains and even improves performance in both Fréchet distance (FD) and diversity (Div) metrics, with our approach consistently outperforming the baseline DSG method. Notably, while DDIM achieves competitive results, DDPM remains superior in most cases. These findings suggest that our method can effectively operate with a reduced number of denoising steps (e.g. $T_r=100$) without significant degradation in quality.

\subsubsection{Temporal acceleration experiments.}
Beyond reducing the number of sampling steps, we leveraged RDLA to accelerate inference in the temporal dimension. By constructing a denoising ladder with step size $l$, we simultaneously denoised $l$ frames at each iteration, achieving an $l$-fold acceleration. Our experiments applied RDLA to DSG, using $N + n^{cont}=108$, ladder step sizes $l \in \{2, 4\}$, and various context sizes $n^{\text{cont}} \in \{8, 18, 28, 38\}$. To maximize speedup, we coupled RDLA with reduced denoising steps, using $T_r \in \{100, 90, 80, 70\}$, corresponding to different $n^{\text{cont}}$ values.

\begin{table}
  \centering
  \begin{tabular}{@{}lcccc@{}}
    \toprule
    Method & $\text{Div}_g \uparrow$ & $\text{Div}_k \uparrow$ & $\text{FD}_g \downarrow$ & $\text{FD}_k \downarrow$ \\
    \midrule
    GT & 272.34 & 213.97 & - & -  \\
    \midrule
    $l=1$, $n^{\text{cont}}=8$ & 256.03 & 179.33 & 3612.14 & 11.74 \\
    \midrule
    $l=2$, $n^{\text{cont}}=8$ & 194.99 & 145.95 & 11001.91 & 28.56 \\
    $l=2$, $n^{\text{cont}}=18$ & 215.41 & 170.92 & 7124.30 & 14.88 \\
    $l=2$, $n^{\text{cont}}=28$ & \bf{222.25} & \bf{173.76} & \bf{5772.40} & \bf{13.65} \\
    $l=2$, $n^{\text{cont}}=38$ & 217.95 & 173.07 & 6728.25 & 20.40 \\
    \midrule
    $l=4$, $n^{\text{cont}}=8$ & 151.90 & 74.22 & 22139.47 & 53.61 \\
    $l=4$, $n^{\text{cont}}=18$ & 137.47 & 75.69 & 20378.92 & 51.63 \\
    $l=4$, $n^{\text{cont}}=28$ & 151.82 & 87.76 & 18250.28 & 45.44 \\
    $l=4$, $n^{\text{cont}}=38$ & \bf{157.77}& \bf{93.22}& \bf{16791.03}& \bf{42.19}\\
    \bottomrule
  \end{tabular}
  \caption{RDLA performance across different ladder steps and context sizes on ZEGGS dataset.}
  \label{tab:rdla_1}
\end{table}

\begin{table}
  \centering
  \begin{tabular}{lcc}
    \toprule
    Method & $\text{MSE}_s \downarrow$ & $\text{MSE}_k \downarrow$ \\
    \midrule
    $l=1$, $n^{\text{cont}}=8$ & \bf{24.64} & \bf{114.01} \\
    \midrule
    $l=2$, $n^{\text{cont}}=8$ & 25.28 & 134.65 \\
    $l=2$, $n^{\text{cont}}=18$ & 25.15 & 139.63 \\
    $l=2$, $n^{\text{cont}}=28$ & $25.33$ & $120.21$ \\
    $l=2$, $n^{\text{cont}}=38$ & $26.02$ & $124.37$ \\
    \midrule
    $l=4$, $n^{\text{cont}}=8$  & $29.63$ & $131.06$ \\
    $l=4$, $n^{\text{cont}}=18$  & 26.04 & 128.37 \\
    $l=4$, $n^{\text{cont}}=28$ & 25.89 & 129.21 \\
    $l=4$, $n^{\text{cont}}=38$ & 25.75 & 126.86 \\
    \bottomrule
  \end{tabular}
  \caption{RDLA performance across different ladder steps and context sizes on ZEGGS dataset using MSE metrics.}
  \label{tab:rdla_2}
\end{table}

Results in \cref{tab:rdla_1} and \cref{tab:rdla_2} indicate that a 2-fold acceleration ($l=2$) leads to a modest drop in metrics, with $n^{\text{cont}}=28$ achieving results comparable to the baseline. However, a 4-fold acceleration ($l=4$) degrades quantitative performance. Despite this, qualitative analysis confirms that motion artifacts are effectively mitigated, and visual quality remains acceptable. Interestingly for larger $l$ wider context improves the performance.

\section{Ablation Study}

To better understand the impact of key design choices in our framework, we conduct an extensive ablation study focusing on different components of our model. Specifically, we investigate the role of noise in context frames and the contributions of individual elements within Rolling Diffusion Ladder Acceleration (RDLA). All experiments are conducted using the DSG backbone on the ZEGGS dataset, $N=100$ and $n^{\text{cont}}=8$ unless otherwise specified.

\subsection{Noise in context frames} \label{sec:noise_in_context}

Our method includes $n^{\text{cont}}$ context frames at the beginning of each rolling window $\mathbf{x}$ of length $N$ to provide a stable conditioning signal. These frames receive minimal noise with $\sigma^2_1=1-\bar{\alpha}^1=0.00004$ during training. To assess the effect of this choice, we train a variant where context frames remain completely noise-free ($\sigma^2_0=0$).

Results in \cref{tab:noise_in_cont_1} and \cref{tab:noise_in_cont_2} indicate that completely removing noise from context frames leads to overfitting, because it required to adjust regularization parameters (the dropout rate increased from $0.2$ to $0.3$, the weight decay increased from $0.01$ to $0.1$) to achieve an acceptable result. But even with this adjustment, performance does not surpass the default setting where context frames contain minimal noise. This suggests that slight corruption of context frames during training improves robustness and generalization.

% As was described in \cref{sec:method}, context frames, i.e. first $n^{\text{cont}}$ frames in a rolling window $\mathbf{x}$ of length $N$, minimum noise level $t_n=1, \forall n \in [0,\ldots,n^{\text{cont}}-1]$ adds to them during training. We found that no adding any noise to the context frames makes model more prone to overfitting and requires increasing of regularization parameters to achieve acceptable performance level. As can be seen from \cref{tab:noise_in_cont_1} our best model trained with zero-noised context frames and increased regularization parameters ($\text{dropout}=0.3$, $\text{weight decay}=0.1$) doesn't outperform model trained as described in \cref{sec:method}. $N=108$, $n^{\text{cont}}=8$. We experimented with DSG based models on ZEGGS dataset.

\begin{table}
  \centering
  \begin{tabular}{@{}lcccc@{}}
    \toprule
    Method & $\text{Div}_g \uparrow$ & $\text{Div}_k \uparrow$ & $\text{FD}_g \downarrow$ & $\text{FD}_k \downarrow$ \\
    \midrule
    $\sigma^2_1=0.00004$& \bf{251.35} & \bf{175.12} & \bf{3831.35} & \bf{8.08} \\
    $\sigma^2_0=0$& 250.17 & 155.25 & 4163.21 & 16.57 \\
    $\sigma^2_0=0$, def set& 234.84 & 136.25 & 8421.93 & 11.90 \\
    \bottomrule
  \end{tabular}
  \caption{Noise in context frames ablation study: $\sigma^2_n=0.00004$ and $\sigma^2_n=0, \forall n \in [0,\ldots,n^{\text{cont}}-1]$ during training, ``def set"" the default setting of the model.}
  \label{tab:noise_in_cont_1}
\end{table}

\begin{table}
  \centering
  \begin{tabular}{lcc}
    \toprule
    Method & $\text{MSE}_s \downarrow$ & $\text{MSE}_k \downarrow$ \\
    \midrule
    $\sigma^2_n=0.00004$ in a context & 24.89 & 121.76 \\
    $\sigma^2_n=0$ & \bf{24.73} & 128.84 \\
    $\sigma^2_n=0$, def set & 25.17 & \bf{116.68} \\
    \bottomrule
  \end{tabular}
  \caption{Noise in context frames ablation study using MSE metrics: $\sigma^2_n=0.00004$ and $\sigma^2_n=0, \forall n \in [0,\ldots,n^{\text{cont}}-1]$ during training, ``def set"" the default setting of the model.}
  \label{tab:noise_in_cont_2}
\end{table}

\subsection{Weighting in loss function}

The weighting function $a(t_n)$ in the training objective in the context of Rolling Diffusion Model for co-speech gestures generation determines the relative importance of different frames in the loss. Prior work suggests that weighting strategies based on signal-to-noise ratio (SNR) can improve training stability in diffusion models. To investigate this, we evaluate a clamped-SNR weighting strategy, which generalizes truncated-SNR and min-SNR weighting:

\begin{equation}\label{eq:weight_snr}
    a(t_n)=\max(\min(\exp(\lambda_{t_n}), \lambda_{\text{max}}), \lambda_{\text{min}})
\end{equation}
where $\lambda_{t_n}=\log(\bar{\alpha}^{t_n}/\sigma^{t_n})$, $\sigma^{t_n}=1 - \bar{\alpha}^{t_n}$ $\forall n \in [0,\ldots,N-1]$. We experiment with different clamping values $\lambda_{\text{min}} \in \{0, 0.001\}$ and $\lambda_{\text{max}} \in \{1, 10\}$.

Results in \cref{tab:weight_snr_1} and \cref{tab:weight_snr_2} show that clamped-SNR weighting does not improve performance over uniform weighting $a(t_n)=1, \forall n$. Instead, uniform weighting consistently yields superior results. These findings contradict the hypothesis that assigning lower weight to higher-noise frames improves training stability. This suggests that equal importance across frames facilitates robust learning for sequential generation.  

\begin{table}
  \centering
  \begin{tabular}{@{}l@{}cccc@{}}
    \toprule
    Method & $\text{Div}_g \uparrow$ & $\text{Div}_k \uparrow$ & $\text{FD}_g \downarrow$ & $\text{FD}_k \downarrow$ \\
    \midrule
    $a(t_n)=1, \forall n$ & \bf{251.35} & \bf{175.12} & \bf{3831.35} & \bf{8.08} \\
    \midrule
    $\lambda_{\text{min}}$, $\lambda_{\text{max}}$ in \cref{eq:weight_snr} \\
    \midrule
    $0.001$,\; $1$ & 205.64 & 120.13 & 9227.13 & 20.21 \\
    $0$,\; $10$ & 211.0 & 109.17 & 9522.56 & 22.45 \\
    \bottomrule
  \end{tabular}
  \caption{Weighting in loss function ablation study: $a(t_n)=1\; \forall n$ and clamped-SNR strategies.}
  \label{tab:weight_snr_1}
\end{table}

\begin{table}
  \centering
  \begin{tabular}{lcc}
    \toprule
    Method & $\text{MSE}_s \downarrow$ & $\text{MSE}_k \downarrow$ \\
    \midrule
    $a(t_n)=1, \forall n$ & 24.89 & 121.76 \\
    \midrule
    $\lambda_{\text{min}}$, $\lambda_{\text{max}}$ in \cref{eq:weight_snr} \\
    \midrule
    $0.001$,\; $1$ & 24.96 & 111.32 \\
    $0$,\; $10$ & \bf{23.82} & \bf{109.48} \\
    \bottomrule
  \end{tabular}
  \caption{Weighting in loss function ablation study using MSE metrics: $a(t_n)=1\; \forall n$ and clamped-SNR strategies.}
  \label{tab:weight_snr_2}
\end{table}

\subsection{RDLA ablations}

To thoroughly evaluate the effectiveness of RDLA, we conduct an ablation study isolating key design components (\cref{tab:rdla_ablation_1} and \cref{tab:rdla_ablation_2}). In all experiments $T=100$. One challenge introduced by RDLA is the potential for abrupt transitions between consecutive denoised blocks, leading to motion discontinuities. To mitigate this, we employ an On-the-Fly Smoothing (OFS) mechanism. As shown in \cref{tab:rdla_ablation_1} and \cref{tab:rdla_ablation_2}, removing OFS results in a slight degradation in motion smoothness. 

RDLA’s inertial loss function $L_{\text{inert}}$ is designed to regularize frame transitions during training. Ablation results in \cref{tab:rdla_ablation_1} and \cref{tab:rdla_ablation_2} show that excluding $L_{\text{inert}}$ increases $\text{FD}_k$ by $6\%$, indicating a loss in motion coherence. This suggests that incorporating an explicit regularization term improves the stability of frame transitions and reduces unwanted jitter in generated gestures.

Given that RDLA introduces structured noise scheduling with a variable step size, we adopt a progressive fine-tuning approach. Initially, the model is trained with a single-frame denoising schedule ($l=1$), followed by a gradual increase in $l$ (e.g., 2, 4, ...). This iterative process enables the model to adapt to larger step sizes while minimizing performance degradation. To evaluate the effectiveness of this strategy, we compare models trained from scratch at $l = 2$ against those fine-tuned from $l = 1$. The results in \cref{tab:rdla_ablation_1} and \cref{tab:rdla_ablation_2} reveal that direct training at $l = 2$ leads to much higher $\text{FD}_g$ and $\text{FD}_k$ values and reduced $\text{Div}_g$ and $\text{Div}_k$, underscoring the advantages of progressive adaptation. Moreover, bypassing training at larger step sizes results in a significant decline in performance, emphasizing the critical role of fine-tuning in preserving fidelity.

\begin{table}
  \centering
  \begin{tabular}{@{}lcccc@{}}
    \toprule
    Method & $\text{Div}_g \uparrow$ & $\text{Div}_k \uparrow$ & $\text{FD}_g \downarrow$ & $\text{FD}_k \downarrow$ \\
    \midrule
    $l=2$, $n^{\text{cont}}=28$ & \bf{222.25} & \bf{173.76} & \bf{5772.40} & \bf{13.65}  \\
    \midrule
    % $+$ OFS, $+$ $L_{\text{inert}}$ & \bf{222.25} & \bf{173.76} & \bf{5772.40} & \bf{13.65} \\
    % $+$ OFS, $-$ $L_{\text{inert}}$ & 220.93 & 174.98 & 6295.34 & 14.20 \\
    $-$ OFS & 219.41 & 168.69 & 6315.91 & 14.57 \\
    $-$ $L_{\text{RDLA}}$ & 219.62 & 166.45 & 6208.70 & 15.73 \\
    $-$ Fine-tuning & 164.45 & 105.89 & 25448.14 & 59.25 \\
    $-$ Training & 170.18 & 86.18 & 19284.87 & 876.38 \\
    \bottomrule
  \end{tabular}
  \caption{RDLA ablation study: impact of On-the-Fly Smoothing (OFS), Inertial loss ($L_{\text{RDLA}}(\theta)$), fine-tuning from $l=1$ and bypassing training for $l=2$, $n^{\text{cont}}=28$. '$-$' indicates non-usage.}
  \label{tab:rdla_ablation_1}
\end{table}

\begin{table}
  \centering
  \begin{tabular}{lcc}
    \toprule
    Method & $\text{MSE}_s \downarrow$ & $\text{MSE}_k \downarrow$ \\
    \midrule
    $l=2$, $n^{\text{cont}}=28$ & 25.33 & \bf{145.18} \\
    \midrule
    $-$ OFS, & \bf{25.30} & 147.46 \\
    $-$ $L_{\text{RDLA}}$ & 25.21 & 148.16 \\
    $-$ Fine-tuning & 25.80 & 321.05  \\
    $-$ Training & 28.16 & 505.69 \\
    \bottomrule
  \end{tabular}
  \caption{RDLA ablation study using MSE metrics: impact of On-the-Fly Smoothing (OFS), Inertial loss ($L_{\text{RDLA}}(\theta)$), fine-tuning from $l=1$ and bypassing training for $l=2$, $n^{\text{cont}}=28$. '$-$' indicates non-usage.}
  \label{tab:rdla_ablation_2}
\end{table}

\section{Qualitative Comparisons}

\begin{figure*}[!t]
  \centering
  \includegraphics[width=\textwidth]{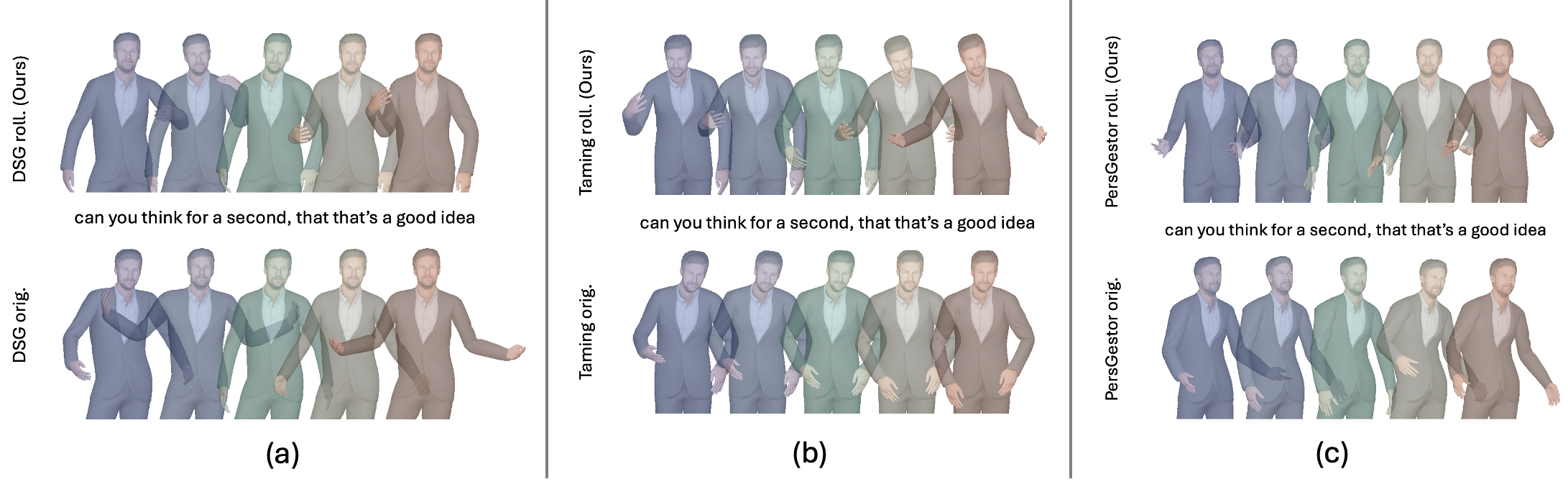}
  \caption{\textbf{Qualitative Comparison on ZEGGS Dataset.} Columns (a–c) show DiffuseStyleGesture, Taming, and PersonaGestor; the baseline is on the bottom row and our version on the top. Our method produces natural, dynamic gestures that are tightly aligned with the speech content. Notably, it uses an energetic hand movement to point toward the interlocutor on the word “you,” effectively emphasizing the rhetorical intent of the phrase “can you think for a second, that that’s a good idea,” while the baselines largely lack emphasis and gesture variation.}
  \label{fig:gestures_vis_1}
\end{figure*}

\begin{figure*}[!t]
  \centering
  \includegraphics[width=\textwidth]{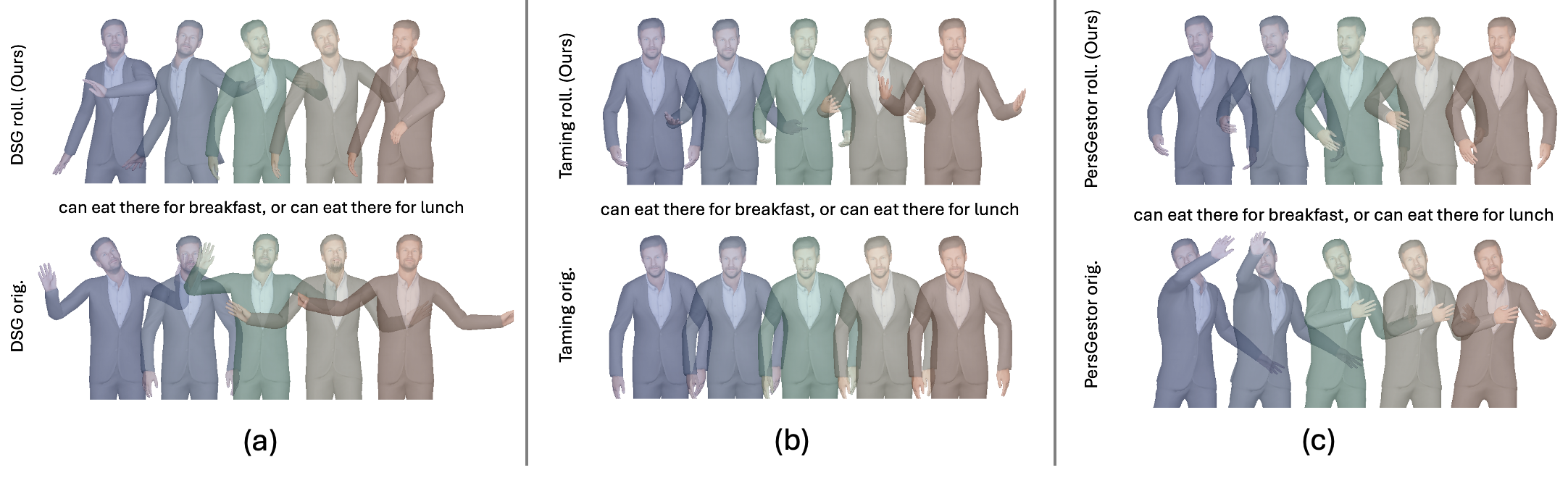}
  \caption{\textbf{Qualitative Comparison on ZEGGS Dataset.} Columns (a–c) show DiffuseStyleGesture, Taming, and PersonaGestor; the baseline is on the bottom row and our version on the top. Our approach generates rhythmically aligned and semantically expressive gestures. In particular, it uses distinct hand movements to accompany and emphasize the repeated structure in “can eat there for breakfast, can eat there for lunch,” effectively conveying a sense of enumeration and emphasis, while the baselines remain mostly static or repetitive.}
  \label{fig:gestures_vis_2}
\end{figure*}

% \bibliography{aaai2026}

% \end{document}


% \end{document}

% \include{ReproducibilityChecklist}
% \input{ReproducibilityChecklist}

\end{document}